\documentclass[sigconf]{acmart}
\usepackage{amsmath,amsfonts,bm}

\def\Figref#1{Figure~\ref{#1}}

\def\secref#1{section~\ref{#1}}
\def\Secref#1{Section~\ref{#1}}
\def\eqref#1{equation~\ref{#1}}
\def\Eqref#1{Equation~\ref{#1}}
\def\1{\bm{1}}

\DeclareMathAlphabet{\mathsfit}{\encodingdefault}{\sfdefault}{m}{sl}
\SetMathAlphabet{\mathsfit}{bold}{\encodingdefault}{\sfdefault}{bx}{n}

\usepackage{subfig}
\usepackage{array,multirow,graphicx}
\usepackage{bm}
\usepackage[normalem]{ulem}

\usepackage{enumitem}

\newcommand{\hide}[1]{} %hide
\newcommand{\vpara}[1]{\vspace{0.1in}\noindent\textbf{#1 }}

\newcommand{\beq}[1]{\small \begin{equation}#1\end{equation}\normalsize}

\newtheorem*{fact}{Fact}
\newtheorem*{hypothesis}{Hypothesis}
\newcommand{\method}{Graph Contrastive Coding}
\newcommand{\shortname}{GCC}

\newcommand{\yd}[1]{\textit{{\color{red} [YD: #1 ]}}}

\AtBeginDocument{%
  \providecommand\BibTeX{{%
    \normalfont B\kern-0.5em{\scshape i\kern-0.25em b}\kern-0.8em\TeX}}}

\setcopyright{acmcopyright}

\copyrightyear{2020}
\acmYear{2020}
\setcopyright{acmcopyright}\acmConference[KDD '20]{Proceedings of the 26th ACM SIGKDD Conference on Knowledge Discovery and Data Mining}{August 23--27, 2020}{Virtual Event, CA, USA}
\acmBooktitle{Proceedings of the 26th ACM SIGKDD Conference on Knowledge Discovery and Data Mining (KDD '20), August 23--27, 2020, Virtual Event, CA, USA}
\acmPrice{15.00}
\acmDOI{10.1145/3394486.3403168}
\acmISBN{978-1-4503-7998-4/20/08}
\begin{document}
\fancyhead{}
\title{\shortname{}: \method{} for 
Graph Neural Network Pre-Training}

\author{Jiezhong Qiu}
\email{qiujz16@mails.tsinghua.edu.cn}
\affiliation{%
  \institution{Tsinghua University}
}

\author{Qibin Chen}
\email{cqb19@mails.tsinghua.edu.cn}
\affiliation{%
  \institution{Tsinghua University}
}

\author{Yuxiao Dong}
\email{yuxdong@microsoft.com}
\affiliation{%
  \institution{Microsoft Research, Redmond}
}

\author{Jing Zhang}
\email{zhang-jing@ruc.edu.cn}
\affiliation{%
  \institution{Remin University}
}

\author{Hongxia Yang}
\email{yang.yhx@alibaba-inc.com}
\affiliation{%
 \institution{DAMO Academy, Alibaba Group}
}

\author{Ming Ding}
\email{dm18@mails.tsinghua.edu.cn}
\affiliation{%
  \institution{Tsinghua University}
}

\author{Kuansan Wang}
\email{kuansan.wang@microsoft.com	}
\affiliation{%
  \institution{Microsoft Research, Redmond}
}

\author{Jie Tang}
\authornote{Jie Tang is the corresponding author.}
\email{jietang@tsinghua.edu.cn}
\affiliation{%
  \institution{Tsinghua University}
}

\renewcommand{\shortauthors}{Qiu et al.}

\begin{abstract}

Graph representation learning has emerged as a powerful technique for addressing real-world problems. 
Various downstream graph learning tasks have benefited from its recent developments, such as node classification, similarity search, and graph classification. %, and link prediction. 
However, prior arts on graph representation learning focus on domain specific problems and train a dedicated model for each graph dataset, which is usually non-transferable to out-of-domain data. 
Inspired by the recent advances in pre-training from natural language processing and computer vision, we design 
Graph Contrastive Coding~(\shortname)\footnote{The code is available at \url{https://github.com/THUDM/GCC}.}---a self-supervised graph neural network pre-training framework---to capture the universal network topological properties across multiple networks. 
We design \shortname's pre-training task as subgraph instance discrimination in and across networks and leverage contrastive learning to empower graph neural networks to learn the intrinsic and transferable structural representations. 
We conduct extensive experiments on three graph learning tasks and ten graph datasets. 
The results show that \shortname{} pre-trained on a collection of diverse datasets can achieve competitive or better performance to its task-specific and trained-from-scratch counterparts. 
This suggests that the \emph{pre-training} and \emph{fine-tuning} paradigm 
presents great potential for graph representation learning. 

\end{abstract}


\begin{CCSXML}
    <ccs2012>
    <concept>
    <concept_id>10002951.10003227.10003351</concept_id>
    <concept_desc>Information systems~Data mining</concept_desc>
    <concept_significance>500</concept_significance>
    </concept>
    <concept>
    <concept_id>10002951.10003260.10003282.10003292</concept_id>
    <concept_desc>Information systems~Social networks</concept_desc>
    <concept_significance>500</concept_significance>
    </concept>
    <concept>
    <concept_id>10010147.10010257.10010258.10010260</concept_id>
    <concept_desc>Computing methodologies~Unsupervised learning</concept_desc>
    <concept_significance>500</concept_significance>
    </concept>
    <concept>
    <concept_id>10010147.10010257.10010293.10010294</concept_id>
    <concept_desc>Computing methodologies~Neural networks</concept_desc>
    <concept_significance>500</concept_significance>
    </concept>
    <concept>
    <concept_id>10010147.10010257.10010293.10010319</concept_id>
    <concept_desc>Computing methodologies~Learning latent representations</concept_desc>
    <concept_significance>500</concept_significance>
    </concept>
    </ccs2012>
\end{CCSXML}
    
\ccsdesc[500]{Information systems~Data mining}
\ccsdesc[500]{Information systems~Social networks}
\ccsdesc[500]{Computing methodologies~Unsupervised learning}
\ccsdesc[500]{Computing methodologies~Neural networks}
\ccsdesc[500]{Computing methodologies~Learning latent representations}

\hide{ %XML
<ccs2012>
<concept>
<concept_id>10002951.10003227.10003351</concept_id>
<concept_desc>Information systems~Data mining</concept_desc>
<concept_significance>500</concept_significance>
</concept>
<concept>
<concept_id>10002951.10003260.10003282.10003292</concept_id>
<concept_desc>Information systems~Social networks</concept_desc>
<concept_significance>500</concept_significance>
</concept>
<concept>
<concept_id>10010147.10010257.10010258.10010260</concept_id>
<concept_desc>Computing methodologies~Unsupervised learning</concept_desc>
<concept_significance>500</concept_significance>
</concept>
<concept>
<concept_id>10010147.10010257.10010293.10010294</concept_id>
<concept_desc>Computing methodologies~Neural networks</concept_desc>
<concept_significance>500</concept_significance>
</concept>
<concept>
<concept_id>10010147.10010257.10010293.10010319</concept_id>
<concept_desc>Computing methodologies~Learning latent representations</concept_desc>
<concept_significance>500</concept_significance>
</concept>
</ccs2012>
}

\keywords{graph representation learning; graph neural networks; graph pre-training; self-supervised learning; contrastive learning}

\maketitle

\section{Introduction}
\label{sec:intro}
\begin{hypothesis}
    Representative graph structural patterns are universal and transferable across networks.
\end{hypothesis}

Over the past two decades, the main focus of network science research has been on discovering and abstracting the universal structural properties underlying different networks. 
For example, Barabasi and Albert show that several types of networks, e.g., World Wide Web, social, and biological networks, have the scale-free property, i.e., all of their degree distributions follow a power law~\cite{albert2002statistical}. 
\citet{leskovec2005graphs} discover that a wide range of real graphs satisfy the densification and shrinking laws. 
Other common patterns across networks include small world~\cite{watts1998collective}, motif distribution~\cite{milo2004superfamilies}, community organization~\cite{newman2006modularity}, and core-periphery structure~\cite{borgatti2000models}, validating our hypothesis at the conceptual level.

In the past few years, however, the paradigm of graph learning has been shifted from structural pattern discovery to graph representation learning~\cite{perozzi2014deepwalk,tang2015line,grover2016node2vec,dong2017metapath2vec,kipf2017semi,qiu2018network,xu2018how,qiu2019netsmf}, motivated by the recent advances in deep learning~\cite{mikolov2013distributed,battaglia2018relational}. 
Specifically, graph representation learning converts the vertices, edges, or subgraphs of a graph into low-dimensional embeddings such that vital structural information of the graph is preserved. 
The learned embeddings from the input graph can be then fed into standard machine learning models for downstream tasks on the same graph.

However, most representation learning work on graphs has thus far focused on learning representations for one single graph or a fixed set of graphs
 and very limited work can be transferred to out-of-domain data and tasks. 
Essentially, those representation learning models aim to learn network-specific structural patterns dedicated for each dataset. 
For example, the DeepWalk embedding model \cite{perozzi2014deepwalk} learned on the Facebook social graph cannot be applied to other graphs. 
In view of (1) this limitation of graph representation learning and (2) the prior arts on common structural pattern discovery, a natural question arises here: \textit{can we universally learn  transferable representative graph embeddings from networks?}

The similar question has also been asked and pursued in natural language processing~\cite{devlin2019bert}, computer vision~\cite{he2019momentum}, and other domains. 
To date, the most powerful solution  is to pre-train a representation learning model from a large dataset, commonly, under the self-supervised setting. % representation learning. 
The idea of pre-training is to use the pre-trained model as a good initialization for fine-tuning over~(different) tasks on unseen datasets. 
For example, BERT~\cite{devlin2019bert} designs language model  pre-training tasks to learn a Transformer encoder~\cite{vaswani2017attention} from 
a large corpus.
The pre-trained Transformer encoder is then adapted to
various NLP tasks~\cite{glue} by fine-tuning.

%\yd{try to add the concept of dist hypo}
%\sout{I want to describe something what I call ``distribution hypothesis'' in social networks. This hypothesis acts as a fundamental assumption in NLP ---  ``words that occur in the same contexts tend to have similar meanings''. I want to say something similar ---  node has the same local structure tend to have similar meaning.The above statement is actually not new. A lot of studies from computer science, sociology and economics, such as strong/weak tie, structural diversity, structure hole, motif, k-core, etc., are actually living examples of the above statement --- they all want to summarize a special structure pattern which could be helpful in understanding the complex effects of networks.}\footnote{\url{https://aclweb.org/aclwiki/Distributional_Hypothesis}}

\vpara{Presented work.}
Inspired by this and the existence of universal graph structural patterns, we study the potential of pre-training representation learning models, specifically, graph neural networks (GNNs),  for graphs. 
Ideally, given a (diverse) set of input graphs, such as the Facebook social graph and the DBLP co-author graph, we aim to pre-train a GNN on them with a self-supervised task, 
and then fine-tune it on different graphs for different graph learning tasks, such as node classification on the US-Airport graph. 
The critical question for GNN pre-training here is: \textit{how to design the pre-training task such that the universal structural patterns in and across networks can be captured and further transferred?}

In this work, we present the \method\ (\shortname) framework to learn structural representations across graphs.  
Conceptually, we leverage the idea of contrastive learning~\cite{wu2018unsupervised} to design the graph pre-training task as instance discrimination. 
Its basic idea is to sample instances from input graphs, treat each of them as a distinct class of its own, and learn to encode and discriminate between these instances. 
%As such, \shortname\ is trained in an self-supervised manner. 
Specifically, there are three questions to answer for \shortname\ such that it can learn the transferable structural patterns: 
\textit{ (1) what are the instances? (2) what are the discrimination rules? and (3) how to encode the instances?}

%In \shortname's pre-training stage, we propose to use a ``subgraph instance discrimination'' pre-training objective. 
In \shortname, we design the pre-training task as \textit{subgraph instance discrimination}. 
%The subgraph instance discrimination task 
Its goal is to distinguish vertices according to their local structures (Cf. \Figref{fig:intro}). 
For each vertex, we sample subgraphs from its multi-hop ego network as instances. 
\shortname{} aims to distinguish between subgraphs sampled from a certain vertex and subgraphs sampled from other vertices. 
Finally, for each subgraph, we use a graph neural network~(specifically, the GIN model~\cite{xu2018how}) as the graph encoder 
to map the underlying structural patterns to latent representations.
As \shortname\ does not assume vertices and subgraphs come from the same graph, the graph encoder is forced to capture universal patterns across different input graphs.  
Given the pre-trained \shortname\ model, we apply it to unseen graphs for addressing downstream tasks. 
%\shortname{} is able to measure the structural similarity between two vertices from different domains, e.g., a user from the Facebook social graph and a scholar from the DBLP co-authorship network, as shown in \Figref{fig:intro}.

To the best of our knowledge, very limited work exists in the field of structural graph representation pre-training to date. 
A very recent one is to design strategies for pre-training GNNs on labeled graphs with node attributes for specific domains~(molecular graphs)~\cite{hu2019pre}. 
Another recent work is InfoGraph~\cite{sun2019infograph}, which focuses 
on learning domain-specific graph-level representations, especially for graph
classification tasks. 
The third related work is by \citet{hu2019workshop}, 
who define several graph learning tasks, such as predicting centrality scores,
to pre-train a GCN~\cite{kipf2017semi} model on synthetic graphs.

We conduct extensive experiments to demonstrate the performance and transferability of \shortname. 
We pre-train the \shortname\ model on a collection of diverse types of graphs and apply the pre-trained model to three downstream graph learning tasks on ten new graph datasets. 
The results suggest that the \shortname\ model  achieves competitive or better results to the state-of-the-art task-specific graph representation learning models that are trained from scratch. 
For example, \textit{for node classification on the US-Airport network,  
\shortname\ pre-trained on the Facebook, IMDB, and DBLP graphs outperforms GraphWave~\cite{donnat2018learning}, ProNE~\cite{zhang2019prone} and Struc2vec~\cite{ribeiro2017struc2vec} which are trained directly on the US-Airport graph, empirically demonstrating our hypothesis at the beginning.} 

{To summarize, our work makes the following four contributions:}
\begin{itemize}[leftmargin=*,itemsep=0pt,parsep=0.5em,topsep=0.3em,partopsep=0.3em]
\item We formalize the problem of graph neural network pre-training across multiple graphs 
and identify its design challenges. 
\item We design the pre-training task as subgraph instance discrimination to capture the universal and  transferable structural patterns from multiple input graphs. 
\item We present the \method\ (\shortname) framework to learn structural graph representations, which leverages contrastive learning to guide the pre-training.  
\item We conduct extensive experiments to demonstrate that for out-of-domain tasks, \shortname\ can offer comparable or superior performance over dedicated graph-specific models.  
\end{itemize}

\begin{figure}
    \centering
    \includegraphics[width=.48\textwidth]{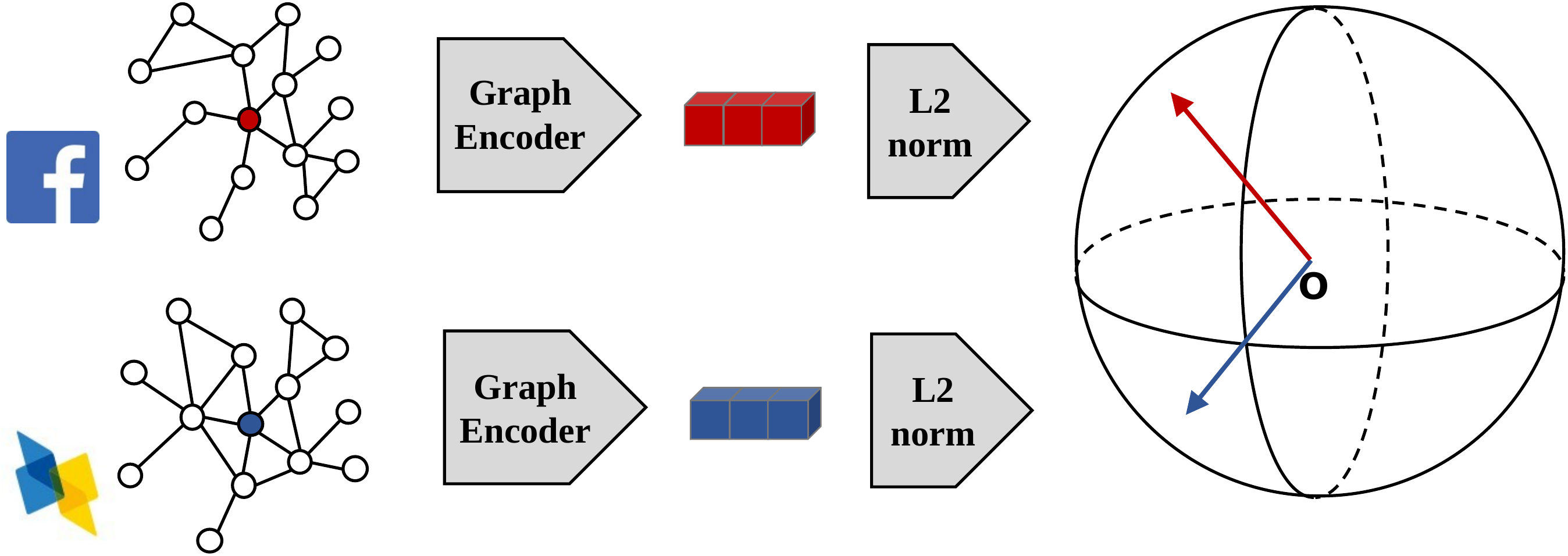}
    \vspace{-0.1in}
    \caption{An illustrative example of \shortname{}. In this example, \shortname{} aims to measure the
    structural similarity between a user from the Facebook
    social graph and a scholar from the DBLP co-authorship network.
    }
    \vspace{-0.1in}
    \label{fig:intro}
\end{figure}

\hide{
\begin{figure}
    \centering
    \includegraphics[width=.48\textwidth]{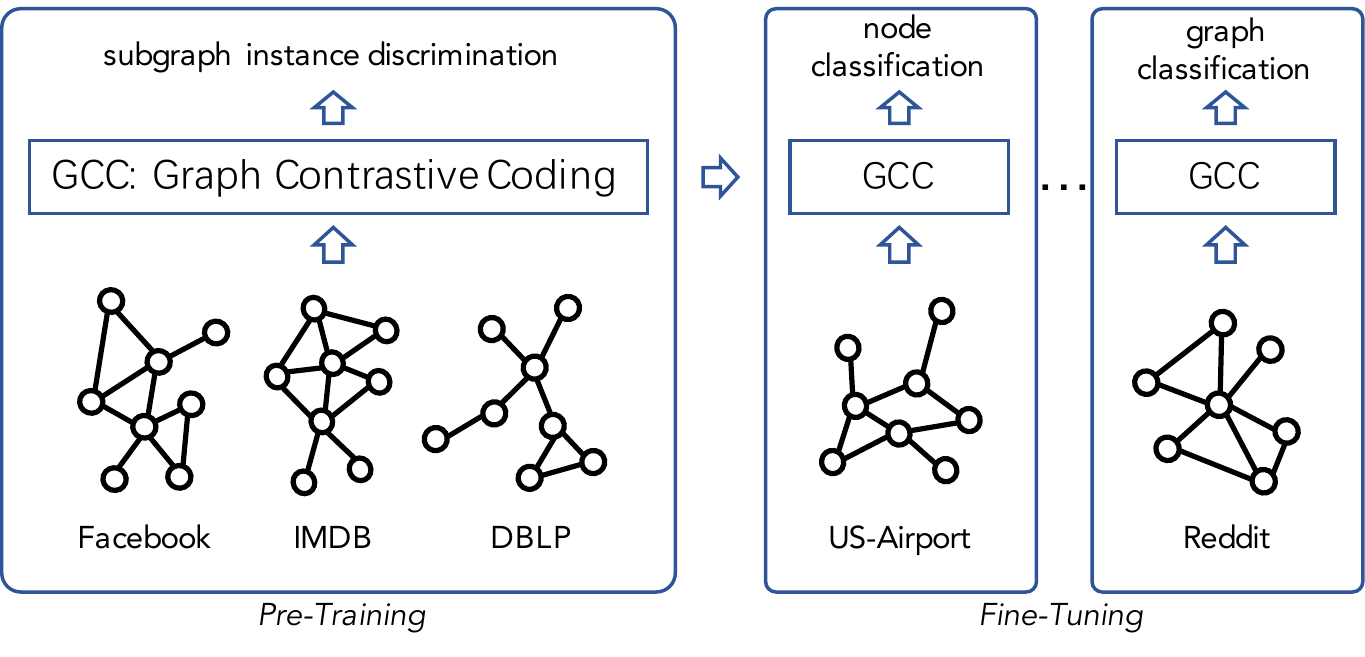}
    \vspace{-0.1in}
    \caption{An illustrative example of \shortname. In this example, \shortname{} aims to measure
    structural similarity between a user from the Facebook
    social graph and a scholar from the DBLP co-authorship network.
    }
    \vspace{-0.1in}
    \label{fig:intro}
\end{figure}

\yd{update a new figure 1. If using the new one, dont forget to change its caption description and the reference text in Introduction}
}

%\clearpage
\section{Related Work}
\label{sec:related}

In this section, we review related work 
of vertex similarity, contrastive learning
and graph pre-training.

\subsection{Vertex Similarity}
\label{subsec:similarity}

Quantifying similarity of vertices in networks/graphs has been 
extensively studied in the past years.
The goal of vertex similarity is to answer questions~\cite{leicht2006vertex} like 
``How similar are these two vertices?''
or
``Which other vertices are most similar to these vertices?''
The definition of similarity can be different 
in different situations. We briefly review 
the following three types of vertex similarity.

\vpara{Neighborhood similarity.}
The basic assumption of neighborhood similarity, a.k.a., proximity,
is that vertices closely connected should be considered similar.
Early neighborhood similarity measures
include Jaccard similarity~(counting common neighbors), 
RWR similarity~\cite{pan2004automatic} and SimRank~\cite{jeh2002simrank},
etc. Most recently developed
network embedding algorithms, such as LINE~\cite{tang2015line}, 
DeepWalk~\cite{perozzi2014deepwalk}, node2vec~\cite{grover2016node2vec},
also follow the neighborhood similarity assumption.

\vpara{Structural similarity.} 
Different from neighborhood similarity which measures
similarity by connectivity, structural similarity doesn't even 
assume vertices are connected.
The basic assumption of structural similarity is that vertices
with similar local structures should be considered similar.
There are two lines of research about modeling structural similarity.
The first line defines 
representative
patterns based on domain knowledge.
Examples include vertex degree,
structural diversity~\cite{ugander2012structural}, 
structural hole~\cite{burt2009structural}, 
k-core~\cite{alvarez2006large}, motif~\cite{milo2002network,benson2016higher}, etc.
Consequently,
models of this genre, such as Struc2vec~\cite{ribeiro2017struc2vec} and RolX~\cite{henderson2012rolx},
usually
involve explicit featurization.
The second line of research leverages the spectral graph theory to model
structural similarity. A recent example is GraphWave~\cite{donnat2018learning}.
In this work, we focus on structural similarity.
Unlike the above two genres,
we adopt contrastive learning and graph neural networks
to learn structural similarity from data.

\vpara{Attribute similarity.} 
Real world graph data always come with rich attributes, 
such as text in citation networks,
demographic information in social networks,
and chemical features in molecular graphs.
Recent graph neural networks models, such as
GCN~\cite{kipf2017semi}, 
GAT~\cite{velivckovic2017graph},
GraphSAGE~\cite{hamilton2017inductive, ying2018graph} 
and MPNN~\cite{gilmer2017neural}, leverage additional attributes as side information or supervised signals
to learn representations which are further used to
 measure vertex similarity.

\subsection{Contrastive Learning}
Contrastive learning is a natural choice to capture
similarity from data.
In natural language processing, 
Word2vec~\cite{mikolov2013distributed} model uses co-occurring words and negative sampling 
to learn word embeddings.
In computer vision, 
a large collection of work~\cite{hadsell2006dimensionality,wu2018unsupervised,he2019momentum,tian2019contrastive} 
learns
self-supervised image representation 
by minimizing the distance between two views of the same image.
In this work, we adopt the InfoNCE loss from \citet{oord2018representation}
and instance discrimination task from \citet{wu2018unsupervised}, as
discussed in \Secref{sec:method}.

\subsection{Graph Pre-Training}

\vpara{Skip-gram based model.}
Early attempts to pre-train graph representations
are skip-gram based network embedding models inspired 
by Word2vec~\cite{mikolov2013distributed}, 
such as LINE~\cite{tang2015line}, 
DeepWalk~\cite{perozzi2014deepwalk}, 
node2vec~\cite{grover2016node2vec}, and metapath2vec~\cite{dong2017metapath2vec}.
Most of them follow the neighborhood similarity assumption, as discussed in \secref{subsec:similarity}.
The representations learned by the above methods are 
tied up with graphs used to train the models, and can not handle out-of-sample problems.
Our \method{}~(\shortname{}) differs from these methods in two aspects.
First, \shortname{} focuses on structural similarity, which is orthogonal to neighborhood similarity.
Second, \shortname{} can be transferred across graphs, even to graphs never seen during pre-training.

\vpara{Pre-training graph neural networks.} 
There are several recent efforts to bring ideas from language pre-training~\cite{devlin2019bert} to 
pre-training graph neural networks~(GNN).
For example, \citet{hu2019pre} pre-train GNN on labeled graphs, especially molecular graphs,
where each vertex~(atom) has an atom type~(such as C, N, O), 
and each edge~(chemical bond) has a bond type~(such as the single bond and double bond).
The pre-training task is to recover atom types and chemical bond types in masked molecular graphs.
Another related work is by \citet{hu2019workshop},
which defines several graph learning tasks to pre-train a GCN~\cite{kipf2017semi}.
Our \shortname{} framework differs from the above methods in two aspects.
First, \shortname{} is for general unlabeled graphs, especially social and information networks.
Second, \shortname{} does not involve explicit featurization and pre-defined graph learning tasks.
%\clearpage

\section{\method\ (\shortname{})}
\label{sec:method}

\begin{figure}
    \centering
    \hspace{-0.05in}
    \includegraphics[width=.48\textwidth]{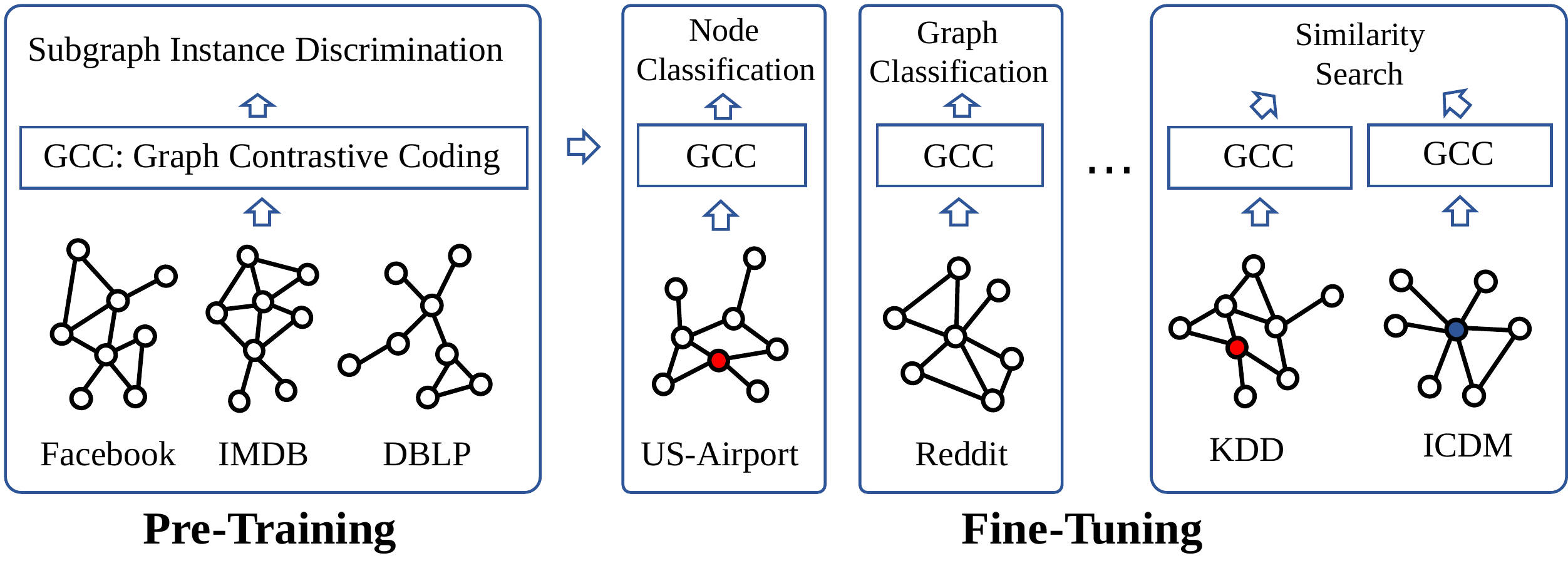}
    \vspace{-0.2in}
    \caption{Overall pre-training and fine-tuning procedures for \method\ (\shortname{}). 
    }
    \vspace{-0.1in}
    \label{fig:framework}
\end{figure}

In this section, we formalize 
the graph neural network~(GNN) pre-training problem. 
%structural graph representation pre-training problem. 
To address it, we present the \method\ (\shortname) framework. 
\Figref{fig:framework} presents the overview of \shortname's pre-training and fine-tuning stages.
%we first briefly review contrastive learning, and then propose our \method{}~(\shortname) framework. 

\subsection{The GNN Pre-Training Problem}

Conceptually, given a collection of graphs from various domains, 
we aim to pre-train a GNN model to capture structural patterns across 
these graphs in a self-supervised manner. 
The model should be able to benefit downstream tasks on different datasets. 
The underlying assumption is that there exist common and transferable structural patterns, such as motifs, across different graphs, as evident in network science literature~\cite{milo2002network,leskovec2005graphs}. 
One illustrative scenario is that we pre-train a GNN model on Facebook, IMDB, and DBLP graphs with self-supervision,  and apply it on the US-Airport network for node classification, as shown in \Figref{fig:framework}. %\dm{This sounds unbelievable, add a sentence for intuition or change the example.}

Formally, the %\textit{structural graph representation pre-training} 
GNN pre-training
problem is to learn a function $f$ that maps a vertex to a low-dimensional feature vector, such that $f$ has the following two properties:
\begin{itemize}[leftmargin=*,itemsep=0pt,parsep=0.5em,topsep=0.3em,partopsep=0.3em]
    \item First, \textit{structural similarity}, it maps vertices with similar local network topologies close to each other in the vector space;
\item Second, \textit{transferability}, it is compatible with vertices and graphs unseen during pre-training.
\end{itemize}

As such, the embedding function $f$ can be adopted in various graph learning tasks, such as 
social role prediction, node classification, and graph classification. 
%In this problem definition, we do not assume $u$ and $v$ to be from the same graph. 

Note that the focus of this work is on structural representation learning without node attributes and node labels, making it different from the common problem setting in graph neural network  research. 
In addition, the goal is to pre-train a structural representation model
%across different types of networks, 
 and apply it to unseen graphs,
differing from traditional network embeddings~\cite{perozzi2014deepwalk,tang2015line, grover2016node2vec,qiu2019netsmf,qiu2018network} and recent attempts on pre-training graph neural networks with attributed graphs as input 
and
applying them 
within a specific domain~\cite{hu2019pre}. 

\subsection{\shortname{} Pre-Training}
 
%To pre-train structural representations on graphs, we present the \shortname{} model. 
Given a set of graphs, our goal is to pre-train a universal graph neural network encoder to capture the  structural patterns behind these graphs.
To achieve this, we need to design 
proper self-supervised tasks and learning objectives for graph structured data.

Inspired by the recent success of contrastive learning in CV~\cite{wu2018unsupervised,he2019momentum} and NLP~\cite{mikolov2013distributed, clark2019electra}, we propose to use \textit{subgraph instance discrimination} as our pre-training task and InfoNCE~\cite{oord2018representation} as our learning objective. 
The  pre-training task  treats each subgraph instance as a distinct class of its own and learns to discriminate between these instances.
%
%We refer to this procedure as \textit{subgraph instance discrimination}.
The promise is that it can output  representations that  captures the similarities between these subgraph instances~\cite{wu2018unsupervised,he2019momentum}. 

From a dictionary look-up perspective, given an encoded query $\bm{q}$ and a dictionary of $K+1$ encoded keys $\{\bm{k}_0, \cdots, \bm{k}_K\}$, contrastive learning looks up a single key~(denoted by $\bm{k}_+$) that $\bm{q}$ matches in the dictionary. 
In this work, we adopt InfoNCE such  that:
\beq{
    \label{eq:obj}
    \mathcal{L}=-\log{\frac{\exp{\left(\bm{q}^\top \bm{k}_+/\tau\right)}}{\sum_{i=0}^K\exp{\left(\bm{q}^\top \bm{k}_i/\tau\right)}}}
}where $\tau$ is the temperature hyper-parameter.
$f_q$ and $f_k$ are two graph neural networks that encode the query instance $x^q$ and each key instance $x^k$ to $d$-dimensional representations, denoted by $\bm{q}=f_q(x^q)$ and $\bm{k}=f_k(x^k)$. 

To instantiate each component in \shortname, we need to answer the following three questions: 
\begin{itemize}[leftmargin=*,itemsep=0pt,parsep=0.5em,topsep=0.3em,partopsep=0.3em]
    \item \textbf{Q1}: How to define subgraph instances in graphs?
    \item \textbf{Q2}: How to define (dis)~similar instance pairs in and across graphs, i.e., for a query $x^q$, which key $x^{k}$ is the matched one? % in the dictionary?
    \item \textbf{Q3}: What are the proper graph encoders $f_q$ and $f_k$?
\end{itemize}

It is worth noting that in our problem setting,  $x^q$ and $x^k$'s are not assumed to be from the same graph.  
%
%\subsection{\shortname\ Design}
%\label{subsec:design}
%
Next we present the design strategies for the \shortname\ pre-training framework by correspondingly answering the aforementioned questions. 

\vpara{Q1: Design (subgraph) instances in graphs.}
The success of contrastive learning framework largely relies on the definition of the data instance. 
It is straightforward for CV and NLP tasks to define an instance as an image or a sentence. However, such ideas cannot be directly
extended to graph data, as instances in graphs are not clearly defined.
Moreover, our pre-training focus is purely on structural representations without additional input features/attributes.
This leaves the natural choice of a single vertex as an instance infeasible, 
as it is not applicable to discriminate between two vertices. 

\begin{figure}
    \centering
    \includegraphics[width=.48\textwidth]{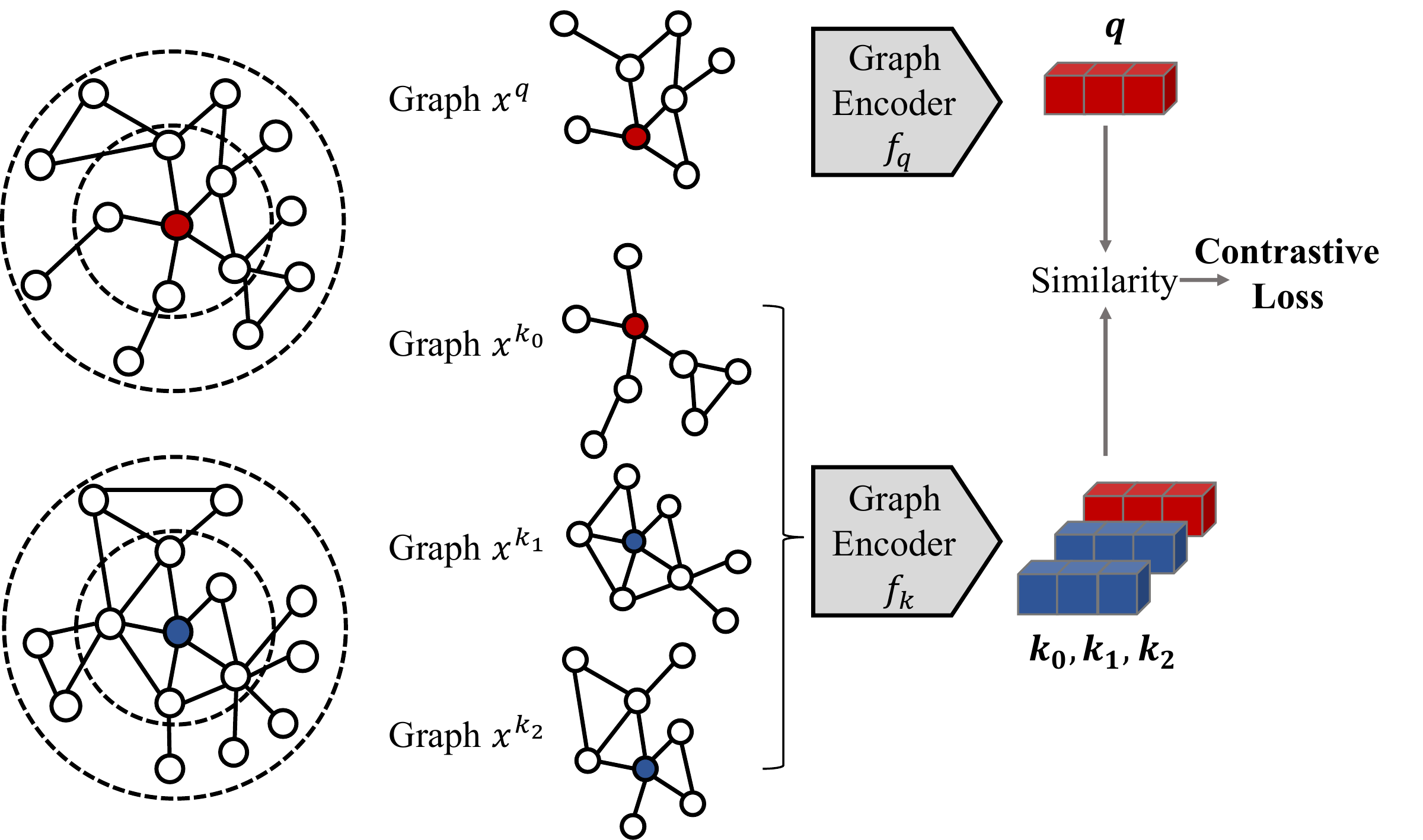}
    \vspace{-0.15in}
    \caption{A running example of \shortname{} pre-training. 
    \hide{
    \small
    {Left: examples of two 2-ego networks with the red and blue vertices as the egos.
    Middle: a similar pair $(x^q, x^{k_0})$ is randomly augmented from the red ego network, and two negative subgraphs, 
    $x^{k_1}$ and $x^{k_2}$, are randomly augmented from another ego network --- the blue one.
    Right: these subgraph instances are encoded by graph neural networks $f_q$ and $f_k$, after which the contrastive loss in \Eqref{eq:obj} encourages a higher similarity score between positive pairs than negative ones. 
    Note that \textit{the two ego networks are not required to come from the same graph.}
    }
    \normalsize
    }
    }
    \vspace{-0.1in}
    \label{fig:story}
\end{figure}

To address this issue, we instead propose to use subgraphs as contrastive instances by extending  
%To address this issue, we propose to 
%extend 
each single vertex to its local structure. 
Specifically, for a certain vertex $v$, we define an instance to be  its $r$-ego network:
\begin{definition}{\textbf{A $r$-ego network.}}
    Let $G=(V, E)$ be a graph, where $V$ denotes
    the set of vertices and $E\subseteq V\times V$ denotes
    the set of edges\footnote{In this work, we consider undirected edges.}.
    For a vertex $v$, its $r$-neighbors are defined as
    $S_v = \{u : d(u,v) \leq r \}$ where $d(u,v)$
    is the shortest path distance between $u$ and $v$ in the graph $G$.
    The $r$-ego network of vertex $v$, denoted by $G_v$, is 
    the  sub-graph induced by  $S_v$.
\end{definition}
The left panel of \Figref{fig:story} shows two examples of 2-ego networks. 
%Given a $r$-ego network, \shortname\ samples different subgraphs and treat them as instances. 
\shortname\ treats each $r$-ego network as a distinct class of its own and encourages the model to distinguish similar instances 
from dissimilar instances. 
Next, we introduce how to define (dis)similar instances. 

\vpara{Q2: Define (dis)similar instances.}% by graph sampling} 
In computer vision~\cite{he2019momentum}, two random data augmentations~(e.g., random crop, random resize, random color jitering, random flip, etc) 
of the same image are treated as a similar instance pair.
In \shortname, we consider two random data augmentations of the same $r$-ego network as a similar instance pair
and define the data augmentation as \textit{graph sampling}~\cite{leskovec2006sampling}. 
Graph sampling is a technique to derive representative subgraph samples from the original graph. 
Suppose we would like to augment vertex $v$'s $r$-ego network ($G_v$), the graph sampling for \shortname\ follows the three steps---random walks with restart~(RWR)~\cite{tong2006fast}, subgraph induction, and anonymization~\cite{micali2016reconstructing, jin2019gralsp}. 
\begin{enumerate}[leftmargin=*,itemsep=0pt,parsep=0.5em,topsep=0.3em,partopsep=0.3em]
    \item \textbf{Random walk with restart.} We start a random walk on $G$ from the ego vertex  $v$.
          The walk iteratively travels to its neighborhood with the probability   proportional to the  edge weight.
          In addition, at each step, with a positive probability the walk returns back to the starting vertex $v$.
    \item \textbf{Subgraph induction.} The random walk with restart collects a subset of vertices surrounding $v$, denoted by $\widetilde{S}_{v}$.
          The sub-graph $\widetilde{G}_v$ induced by $\widetilde{S}_v$ is then regarded as an augmented version
          of the $r$-ego network $G_v$. This step is also known as the Induced Subgraph Random Walk Sampling~(ISRW).
    \item \textbf{Anonymization.} We anonymize the sampled graph $\widetilde{G}_v$ by re-labeling its vertices to be $\{1,2, \cdots, |\widetilde{S}_{v}|\}$, in arbitrary order\footnote{Vertex order doesn't matter because most of graph neural networks are invariant to permutations of their inputs~\cite{battaglia2018relational}.}.
\end{enumerate}
We repeat the aforementioned procedure twice to create two data augmentations, which form a similar instance pair $(x^q$, $x^{k_+})$.
If two subgraphs are augmented from different $r$-ego networks, 
we treat them as a dissimilar instance pair ($x^q$, $x^{k}$) with $k\neq k_+$. 
It is worth noting that all the above graph operations---random walk with restart, subgraph induction, and anonymization---are available in the DGL package~\cite{wang2019deep}. 

\textbf{Discussion on graph sampling.} 
In random walk with restart sampling, the restart probability controls the radius of ego-network (i.e., $r$) which \shortname\ conducts data augmentation on. In this work, 
we follow \citet{qiu2018deepinf} to use 0.8 as the restart probability.
The proposed \shortname{} framework is flexible to other graph sampling algorithms, such as 
neighborhood sampling~\cite{hamilton2017inductive} and forest fire~\cite{leskovec2006sampling}. 

\textbf{Discussion on anonymization.} 
Now we discuss the intuition behind the anonymization step in the above procedure. 
This step is designed to keep the underlying structural patterns and hide the exact vertex indices.
This design  avoids learning a trivial solution to subgraph instance discrimination, i.e., simply checking whether vertex indices of two subgraphs match. 
Moreover, it facilitates the transfer of the learned model across different graphs as such a model is not associated with a particular vertex set.

\vpara{Q3: Define graph encoders.}
Given two sampled subgraphs $x^q$ and $x^k$, 
\shortname{} encodes them via two graph neural network encoders $f_q$ and $f_k$, respectively.
%The encoded $d$-dimensional vectors are then used for calculating the similarity between them in the vector space. 
Technically, any graph neural networks~\cite{battaglia2018relational} can be used here as the encoder, and the \shortname\ model is not sensitive to different choices. 
In practice, we adopt the Graph Isomorphism Network~(GIN)~\cite{xu2018how}, a state-of-the-art graph neural network model, as our graph encoder.
%which maps the input graphs to $d$-dimensional vectors. 
%
Recall that we focus on structural representation pre-training  while most GNN models require vertex features/attributes as input. 
To bridge the gap, we propose to leverage the graph structure of each sampled subgraph to initialize vertex features. 
Specifically, we define the \emph{generalized positional embedding} as follows:
\begin{definition}{\textbf{Generalized positional embedding.}} 
    For each subgraph, %we first form its normalized graph Laplacian and then compute its top eigenvectors as its positional embedding. 
    its generalized positional embedding is defined to be the top eigenvectors of its normalized graph Laplacian.
    Formally, suppose one subgraph has adjacency matrix $\bm{A}$ and degree matrix $\bm{D}$, we conduct eigen-decomposition on its normalized graph Laplacian s.t.
    $\bm{I}-\bm{D}^{-1/2} \bm{A} \bm{D}^{-1/2} = \bm{U} \bm{\Lambda} \bm{U}^\top$, where the top eigenvectors in $\bm{U}$~\cite{von2007tutorial} are defined as generalized positional embedding. 
\end{definition}
The generalized positional embedding is inspired by the Transformer model in NLP~\cite{vaswani2017attention}, where the sine and cosine functions of different frequencies are used to define the positional embeddings in word sequences. 
Such a definition is deeply connected with graph Laplacian as follows. 
\begin{fact}
    The Laplacian of path graph has eigenvectors:
    $
        \bm{u}_k(i) = \cos{(\pi k i/n - \pi k / 2n)}
    $, for $1\leq k \leq n, 1\leq i \leq n$.
    Here $n$ is the number of vertices in the path graph, and $\bm{u}_k(i)$ is the entry at $i$-th row and $k$-the column of $\bm{U}$, i.e., $\bm{U} = \begin{bmatrix} \bm{u}_1 & \cdots& \bm{u}_n\end{bmatrix}$.
\end{fact}
The above fact shows that the positional embedding in sequence models can be viewed as
Laplacian eigenvectors of path graphs. 
This inspires us to generalize the positional embedding from path graphs to arbitrary graphs.
The reason for using the normalized graph Laplacian rather than the unnormalized version is that 
path graph is a regular graph~(i.e., with constant degrees) while real-world graphs are often irregular and have skewed degree distributions.
%We name it to be generalized positional embedding.
In addition to the generalized positional embedding, we also add the one-hot encoding of vertex degrees~\cite{xu2018how} and
the binary indicator of the ego vertex~\cite{qiu2018deepinf} as vertex features. 
After encoded by the graph encoder, the final $d$-dimensional output vectors are then normalized by their L2-Norm~\cite{he2019momentum}.

\vpara{A running example.} 
We illustrate a running example of \shortname{} pre-training in \Figref{fig:story}. 
%\Figref{subfig:ego} shows an example of 2-ego network.
%Given a 2-ego network on the left, 
For simplicity, we set the dictionary size to be 3. %, i.e., $K=2$. 
\shortname{} first randomly augments two subgraphs $x^q$ and $x^{k_0}$ from a 2-ego network on the left panel of \Figref{fig:story}. 
Meanwhile,  another two subgraphs, $x^{k_1}$ and $x^{k_2}$, are generated from a noise distribution --- 
in this example, they are randomly augmented from another 2-ego network on the left panel of \Figref{fig:story}. 
Then the two graph encoders,  $f_q$ and $f_k$, map the query and the three keys to  low-dimensional vectors --- $\bm{q}$ and  $\{\bm{k}_0, \bm{k}_1, \bm{k}_2\}$. 
Finally, the contrastive loss in Eq.~\ref{eq:obj} encourages the model to recognize $(x^q, x^{k_0})$
as a similar instance pair and distinguish them from dissimilar instances, i.e., $\{x^{k_1}, x^{k_2}\}$.

\vpara{Learning.} In contrastive learning, it is required to maintain the $K$-size dictionary and encoders. 
Ideally, %to compute the contrastive loss in Eq. \ref{eq:obj},
in Eq. \ref{eq:obj},
the dictionary should cover 
as many instances as possible, making $K$ extremely large.  
However, due to the computational constraints, we usually design and adopt economical strategies to 
effectively build and maintain the dictionary, such as {end-to-end}~(E2E) and {momentum contrast}~(MoCo)~\cite{he2019momentum}. 
We discuss the two strategies as follows.

E2E samples mini-batches of instances and  considers samples in the same mini-batch as the dictionary.
The objective in Eq.~\ref{eq:obj} is then optimized with respect to parameters of both $f_q$ and $f_k$, both of which  can accept gradient updates by backpropagation consistently. 
The main drawback of E2E is that the dictionary size is constrained by the batch size. %and computation resources such as GPU memory. 

MoCo is designed to increase the dictionary size without additional backpropagation costs.
Concretely, MoCo maintains a queue of samples from preceding mini-batches. 
During optimization, MoCo only updates the parameters of $f_q$~(denoted by $\theta_q$) by backpropagation. 
The parameters of $f_k$~(denoted by $\theta_k$) are not updated by gradient descent.
\citet{he2019momentum} propose a momentum-based update rule for $\theta_k$.  
Formally,
MoCo updates $\theta_k$  by 
$\theta_k \leftarrow m \theta_k + (1-m) \theta_q$, where $m \in [0, 1)$ is a momentum hyper-parameter.
The above momentum update rule gradually propagates the update in $\theta_q$ to $\theta_k$, 
making $\theta_k$ evolve smoothly and consistently. 
In summary, MoCo achieves a larger dictionary size at the expense of dictionary consistency, i.e., 
the key representations in the dictionary are encoded by a smoothly-varying key encoder.

In addition to E2E and MoCo, there are other contrastive learning mechanisms to maintain the dictionary, such as memory bank~\cite{wu2018unsupervised}. 
Recently, \citet{he2019momentum} show that MoCo is a more effective option than memory bank in computer vision tasks. 
Therefore, we mainly focus on E2E and MoCo for \shortname.

\subsection{\shortname{} Fine-Tuning}
\label{subsec:ft}

\vpara{Downstream tasks.} Downstream tasks 
in graph learning generally fall into two categories---graph-level and node-level,
where the target is to predict labels of graphs or nodes, respectively. 
For graph-level tasks, 
the input graph itself can be encoded by \shortname{} to achieve the representation.
For node-level tasks, 
the node representation can be defined by encoding its $r$-ego networks~(or subgraphs augmented from its $r$-ego network).
In either case, the encoded representations are then fed into downstream tasks to predict task-specific outputs.

\vpara{Freezing vs. full fine-tuning.}
\shortname{} offers two fine-tuning strategies for downstream tasks---the freezing mode and full fine-tuning mode.
In the freezing mode, we freeze the parameters of the pre-trained graph encoder $f_q$ and treat it as a static feature extractor, then the classifiers catering for specific downstream tasks are trained on top of the extracted features. 
%The freezing mode fits for the situation that many tasks are performed on a new graph.
In the full fine-tuning mode, the graph encoder $f_q$ initialized with pre-trained parameters is trained end-to-end together with the classifier on a downstream task.
More implementation details about fine-tuning are available in \Secref{subsec:exp_transfer}.

\vpara{\shortname{} as a local algorithm.}
As a graph algorithm, \shortname{} belongs to the local algorithm category~\cite{spielman2013local,teng2016scalable}, in which the algorithms only involve local explorations of the input (large-scale) network, 
since
\shortname{} explores 
local structures by random walk based graph sampling methods. 
Such a property enables \shortname{} to scale to large-scale graph learning tasks and to be friendly to the distributed computing setting.

%\clearpage

\section{Experiments}
\label{sec:exp}

In this section, we evaluate \shortname{}
on three graph learning tasks---node classification, graph classification, and similarity search,
which have been commonly
used to benchmark
graph learning algorithms~\cite{yanardag2015deep,ribeiro2017struc2vec,donnat2018learning,xu2018how,sun2019infograph}.
We first introduce the self-supervised pre-training settings in \Secref{subsec:exp_pretrain},
and then report \shortname{} fine-tuning results on those three graph learning tasks in \Secref{subsec:exp_transfer}.

\subsection{Pre-Training}
\label{subsec:exp_pretrain}

\begin{table*}[htbp]
    \caption{Datasets for pre-training, sorted by number of vertices.}
    \vspace{-0.1in}
    \centering
    \small
    \begin{tabular}{l|rrrrrr}
        \toprule
        Dataset & Academia & DBLP~(SNAP) & DBLP~(NetRep) & IMDB      & Facebook   & LiveJournal \\\midrule
        $|V|$   & 137,969  & 317,080     & 540,486       & 896,305   & 3,097,165  & 4,843,953   \\
        $|E|$   & 739,384  & 2,099,732   & 30,491,458    & 7,564,894 & 47,334,788 & 85,691,368  \\\bottomrule
    \end{tabular}
    \label{tbl:pretrain_dataset}
    \normalsize
    \vspace{-0.1in}
\end{table*}

\vpara{Datasets.}
Our self-supervised pre-training is performed on six graph datasets, which can be categorized into two groups---\textit{academic graphs} and \textit{social graphs}.
As for academic graphs, we collect
the  Academia dataset from NetRep~\cite{ritchie2016scalable} as well as
two DBLP datasets
from SNAP~\cite{yang2015defining} and NetRep~\cite{ritchie2016scalable}, respectively.
As for social graphs, we collect Facebook and IMDB datasets from NetRep~\cite{ritchie2016scalable},
as well as a LiveJournal dataset from SNAP~\cite{backstrom2006group}.
Table~\ref{tbl:pretrain_dataset} presents
the detailed statistics of datasets for pre-training.

\vpara{Pre-training settings.}
We train for 75,000 steps and use Adam~\cite{kingma2014adam} for optimization with learning rate of 0.005,
$\beta_{1}=0.9, \beta_{2}=0.999, \epsilon=1 \times 10^{-8}$, weight decay of 1e-4, learning rate warmup over the first $7,500$ steps,
and linear decay of the learning rate after $7,500$ steps.
Gradient norm clipping is applied with range $[-1,1]$.
For MoCo, we use mini-batch size of $32$, dictionary size of $16,384$, and momentum  $m$ of $0.999$.
For E2E, we use mini-batch size of $1,024$.
For both MoCo and E2E, the temperature $\tau$ is set as $0.07$, and
 we adopt GIN~\cite{xu2018how} with 5 layers and 64 hidden units each layer as our encoders.
Detailed hyper-parameters can be found in Table~\ref{tbl:hyper} in the Appendix.

\subsection{Downstream Task Evaluation}
\label{subsec:exp_transfer}

In this section, we apply \shortname{} to three graph learning tasks
including node classification, graph classification, and similarity search.
%We compare \shortname{} with various baselines.
As prerequisites, we discuss the two fine-tuning strategies of \shortname{} as well as the baselines we compare with.

\vpara{Fine-tuning.} As we discussed in \Secref{subsec:ft}, we adopt two fine-tuning strategies for \shortname{}.
We select logistic regression or SVM from the scikit-learn~\cite{pedregosa2011scikit} package as the linear classifier for the freezing strategy\footnote{
    In node classification tasks, we follow Struc2vec to use logistic regression.
    For graph classification tasks, we follow DGK~\cite{yanardag2015deep} and GIN~\cite{xu2018how} to use SVM~\cite{chang2011libsvm}.}.
As for the full fine-tuning strategy,
we use the Adam optimizer with learning rate 0.005, learning rate warmup over the first 3 epochs, and linear  learning rate decay after 3 epochs.

\vpara{Baselines.} Baselines can be categorized
into two categories.
In the first category, the baseline models learn vertex/graph representations from unlabeled graph data and then feed them into logistic regression or SVM.
Examples include DGK~\cite{yanardag2015deep}, Struc2vec~\cite{ribeiro2017struc2vec},
GraphWave~\cite{donnat2018learning},
graph2vec\cite{narayanan2017graph2vec} and
InfoGraph~\cite{sun2019infograph}.
\shortname{} with the freezing setting belongs to this category. 
In the second category,
the models are optimized in an end-to-end supervised manner.
Examples include DGCNN~\cite{zhang2018end} and GIN~\cite{xu2018how}.
\shortname{} with the full fine-tuning setting belongs to this category.

For a fair comparison, we fix the representation dimension of all models
to be 64 except graph2vec and InfoGraph\footnote{We allow
    them to use their preferred dimension size in their papers: graph2vec uses 1024 and InfoGraph uses 512.}. 
    The details of baselines  will be discussed later.

\subsubsection{Node Classification}\hfill

\vpara{Setup.}
The node classification task is to predict unknown node labels in a partially labeled network.
To evaluate \shortname{},
we sample a subgraph centered at each vertex and apply \shortname{} on it.
Then the obtained representation is fed into an output layer to predict the node label.
As for datasets, we adopt US-Airport~\cite{ribeiro2017struc2vec} and H-index~\cite{zhang2019oag}.
US-Airport consists of the airline activity data among 1,190 airports.
The 4 classes indicate different activity levels of the airports.
H-index is a co-authorship graph extracted from OAG~\cite{zhang2019oag}.
The labels indicate whether the h-index of the author is above or below the median.

\vpara{Experimental results.}
We compare \shortname{} with
ProNE~\cite{zhang2019prone},
GraphWave~\cite{donnat2018learning},
and Struc2vec~\cite{ribeiro2017struc2vec}.
Table~\ref{tbl:node} represents the results.
It is worth noting that,
under the freezing setting,
the graph encoder in \shortname{} is not trained on either
US-Airport or H-Index dataset,
which
other baselines use as training data.
This places \shortname{} at a disadvantage.
However, \shortname{}~(MoCo, freeze) performs competitively
to Struc2vec
in US-Airport, and achieves the best performance in H-index where Struc2vec cannot finish in one day.
Moreover,
\shortname{} can be further boosted
by fully fine-tuning on the target US-Airport or H-Index domain.

\begin{table}[htbp]
    \caption{Node classification.
        \hide{
        \small 
        ``E2E'' and ``MoCo'' are two different contrastive learning mechanisms.
        ``freeze'' and ``full'' indicate 
        fine-tuning with pre-trained graph encoder frozen and fully fine-tuning all parameters, respectively.
        \normalsize
        }
    }
    \vspace{-0.1in}
    \label{tbl:node}
    \small
    \centering
    \begin{tabular}{l|rrrrrr}
        \toprule
        Datasets                    & US-Airport    & H-index       \\\midrule
        $|V|$                       & 1,190         & 5,000         \\
        $|E|$                       & 13,599        & 44,020        \\\midrule
        ProNE                       & 62.3          & 69.1          \\
        GraphWave                   & 60.2          & 70.3          \\
        Struc2vec                   & \textbf{66.2} & > 1 Day       \\
        \shortname{}~(E2E, freeze)  & 64.8          & \textbf{78.3} \\
        \shortname{}~(MoCo, freeze) & 65.6          & 75.2          \\\midrule
        %GraphSAGE           &               &               &               &               \\
        \shortname~(rand, full)     & 64.2          & 76.9          \\
        \shortname~(E2E, full)      & \textbf{68.3} & 80.5          \\
        \shortname~(MoCo, full)     & 67.2          & \textbf{80.6} \\\bottomrule
    \end{tabular}
    \normalsize
    \vspace{-0.1in}
\end{table}

\subsubsection{Graph Classification}\hfill

\begin{table}[htbp]
    \caption{Graph classification.}
    \vspace{-0.1in}
    \label{tbl:graph_classification}
    \centering
    \small
     \setlength{\tabcolsep}{0.8mm}{

    \begin{tabular}{@{~~}l|@{~~}rrrrrr@{~~}}
        \toprule
        Datasets                    & IMDB-B        & IMDB-M        & COLLAB        & RDT-B         & RDT-M         \\\midrule
        \# graphs                   & 1,000         & 1,500         & 5,000         & 2,000         & 5,000         \\
        \# classes                  & 2             & 3             & 3             & 2             & 5             \\
        Avg. \# nodes               & 19.8          & 13.0          & 74.5          & 429.6         & 508.5         \\\midrule
        DGK                         & 67.0          & 44.6          & 73.1          & 78.0          & 41.3          \\
        graph2vec                   & 71.1          & 50.4          & --            & 75.8          & 47.9          \\
        InfoGraph                   & \textbf{73.0} & \textbf{49.7} & --            & 82.5          & 53.5          \\
        \shortname{}~(E2E, freeze)  & 71.7          & 49.3          & 74.7          & 87.5          & 52.6          \\
        \shortname{}~(MoCo, freeze) & 72.0          & 49.4          & \textbf{78.9} & \textbf{89.8} & \textbf{53.7} \\\midrule
        DGCNN                       & 70.0          & 47.8          & 73.7          & --            & --            \\
        GIN                         & \textbf{75.6} & \textbf{51.5} & 80.2          & \textbf{89.4} & \textbf{54.5} \\
        \shortname{}~(rand, full)   & \textbf{75.6} & 50.9          & 79.4          & 87.8          & 52.1          \\
        \shortname{}~(E2E, full)    & 70.8          & 48.5          & 79.0          & 86.4          & 47.4          \\
        \shortname{}~(MoCo, full)   & 73.8          & 50.3          & \textbf{81.1} & 87.6          & 53.0          \\
        \bottomrule
    \end{tabular}}
    \normalsize
    \vspace{-0.1in}
\end{table}

\vpara{Setup.}
We use five datasets from \citet{yanardag2015deep}---COLLAB, IMDB-BINARY, IMDB-MULTI, REDDITBINARY and REDDIT-MULTI5K, which are
widely benchmarked in recent graph classification models~\cite{hu2019pre,sun2019infograph, zhang2018end}.
Each dataset is a set of graphs where each graph is associated with a label.
To evaluate \shortname{} on this task, we use raw input graphs as the input of \shortname{}. 
Then the encoded graph-level representation is fed into a classification layer to predict the label of the graph.
We compare \shortname{} with several recent developed graph classification models,
including Deep Graph Kernel~(DGK)~\cite{yanardag2015deep}, graph2vec~\cite{narayanan2017graph2vec},
InfoGraph~\cite{sun2019infograph}, DGCNN~\cite{zhang2018end} and
GIN~\cite{xu2018how}.
Among these baselines, DGK, graph2vec and InfoGraph belong to the first category,
while DGCNN and GIN belong to the second category.

\vpara{Experimental results.}
Table~\ref{tbl:graph_classification} shows the comparison.
In the first category,
\shortname{}~(MoCo, freeze)  performs competitively
to InfoGraph in IMDB-B and IMDB-M, while achieves
the best performance in other datasets.
Again, we want to emphasize that DGK, graph2vec and InfoGraph all need to be pre-trained
on target domain graphs, but \shortname{} only relies on the graphs listed in Table~\ref{tbl:pretrain_dataset} for pre-training.
In the second category, we compare \shortname{} with DGCNN and GIN.
\shortname{} achieves better performance than DGCNN and comparable performance to GIN.
GIN is a recently proposed SOTA model for graph classification.
We follow the instructions in the paper~\cite{xu2018how}
to train GIN and report the detailed results in Table~\ref{tbl:gin} in the Appendix.
We can see that, in each dataset, the best performance
of GIN is achieved by different hyper-parameters. And by varying
hyper-parameters, GIN's performance could be sensitive.
However,
\shortname{} on all datasets shares the same pre-training/fine-tuning
hyper-parameters,
showing its robustness on graph classification.

\label{subsubsec:similarity}

\begin{table}[htbp]
    \caption{Top-$k$ similarity search~($k=20, 40$).}
    \vspace{-0.1in}
    \centering
    \small
    \begin{tabular}{l | r r |r r| r r@{~}}
        \toprule
                            & \multicolumn{2}{c|}{KDD-ICDM} & \multicolumn{2}{c|}{SIGIR-CIKM} & \multicolumn{2}{c}{SIGMOD-ICDE}                                                       \\\midrule
        $|V|$               & 2,867                         & 2,607                           & 2,851                           & 3,548           & 2,616           & 2,559           \\
        $|E|$               & 7,637                         & 4,774                           & 6,354                           & 7,076           & 8,304           & 6,668           \\
        \#~groud truth      & \multicolumn{2}{r|}{697}      & \multicolumn{2}{r|}{874}        & \multicolumn{2}{r}{898}                                                               \\\midrule
        $k$                 & 20                            & 40                              & 20                              & 40              & 20              & 40              \\            \midrule
        Random              & 0.0198                        & 0.0566                          & 0.0223                          & 0.0447          & 0.0221          & 0.0521          \\
        RolX                & 0.0779                        & 0.1288                          & 0.0548                          & 0.0984          & 0.0776          & 0.1309          \\
        Panther++           & 0.0892                        & 0.1558                          & \textbf{0.0782}                 & 0.1185          & 0.0921          & 0.1320          \\
        GraphWave           & 0.0846                        & \textbf{0.1693}                 & 0.0549                          & 0.0995          & \textbf{0.0947} & \textbf{0.1470} \\
        \shortname{}~(E2E)  & \textbf{0.1047}               & 0.1564                          & 0.0549                          & \textbf{0.1247} & 0.0835          & 0.1336          \\
        \shortname{}~(MoCo) & 0.0904                        & 0.1521                          & 0.0652                          & 0.1178          & 0.0846          & 0.1425          \\
        \bottomrule
    \end{tabular}
    \normalsize
    \label{tbl:similarity_search_dataset}
    \vspace{-0.1in}
\end{table}

\subsubsection{Top-$k$ Similarity Search}\hfill

\vpara{Setup.} We adopt the co-author dataset from
\citet{zhang2015panther}, which are the conference co-author graphs of
KDD, ICDM, SIGIR, CIKM, SIGMOD, and ICDE.
The problem of top-$k$ similarity search is defined as follows.
Given two graphs $G_1$ and $G_2$, for example
KDD and ICDM co-author graphs,
we want to find the most similar vertex $v$ from
$G_1$ for each vertex $u$ in $G_2$. In this dataset, the ground truth is defined to be authors publish in both conferences.
Note that similarity search
is an unsupervised task, so we evaluate \shortname{} without fine-tuning.
Especially, we first extract two subgraphs centered at
$u$ and $v$ by random walk with restart graph sampling.
After encoding them by \shortname{}, we measure the similarity
score between $u$ and $v$ to be the inner product of their representations.
Finally,  by sorting the above scores, we use HITS@10~(top-10 accuracy) to measure the performance of different methods.
We compare \shortname{} with
RolX~\cite{henderson2012rolx}, Panther++~\cite{zhang2015panther} and GraphWave~\cite{donnat2018learning}.
We also provide random guess results for reference.

\vpara{Experimental results.}
Table~\ref{tbl:similarity_search_dataset} presents the performance of
different methods on top-$k$ similarity search task
in three co-author networks.
We can see that, compared with Panther++~\cite{zhang2015panther}
and GraphWave~\cite{donnat2018learning} which are trained in place
on co-author graphs,
simply applying pre-trained \shortname{} can be
competitive.

\emph{
    Overall,
    we show that
    a graph neural network encoder pre-trained on several popular graph datasets
    can be directly adapted
    to new graph datasets and  unseen graph learning tasks.
    More importantly,
    compared with models trained from scratch,
    the reused model
    achieves competitive and sometimes better
    performance.
    This demonstrates the
    transferability of graph structural patterns and
    the effectiveness of our \shortname{} framework in
    capturing these patterns.
}

\subsection{Ablation Studies}

\vpara{Effect of pre-training.}
It is still not clear if \shortname{}'s good performance 
is due to pre-training or the expression power of its GIN~\cite{xu2018how} encoder.
To answer this question, we fully fine-tune \shortname{} with
its GIN encoder randomly initialized, which is equivalent to train a GIN encoder from scratch.
We name this model \shortname{}~(rand), as shown in
Table~\ref{tbl:node} and Table~\ref{tbl:graph_classification}.
In all datasets except IMDB-B, \shortname{}~(MoCo) outperforms its randomly initialized counterpart, showing that pre-training always provides a better start point for fine-tuning than
random initialization. For IMDB-B, we attribute it to the domain shift between pre-training data and down-stream tasks.

\begin{figure}[t]
    \centering
    \subfloat[US-Airport]{
        \includegraphics[width=0.23\textwidth]{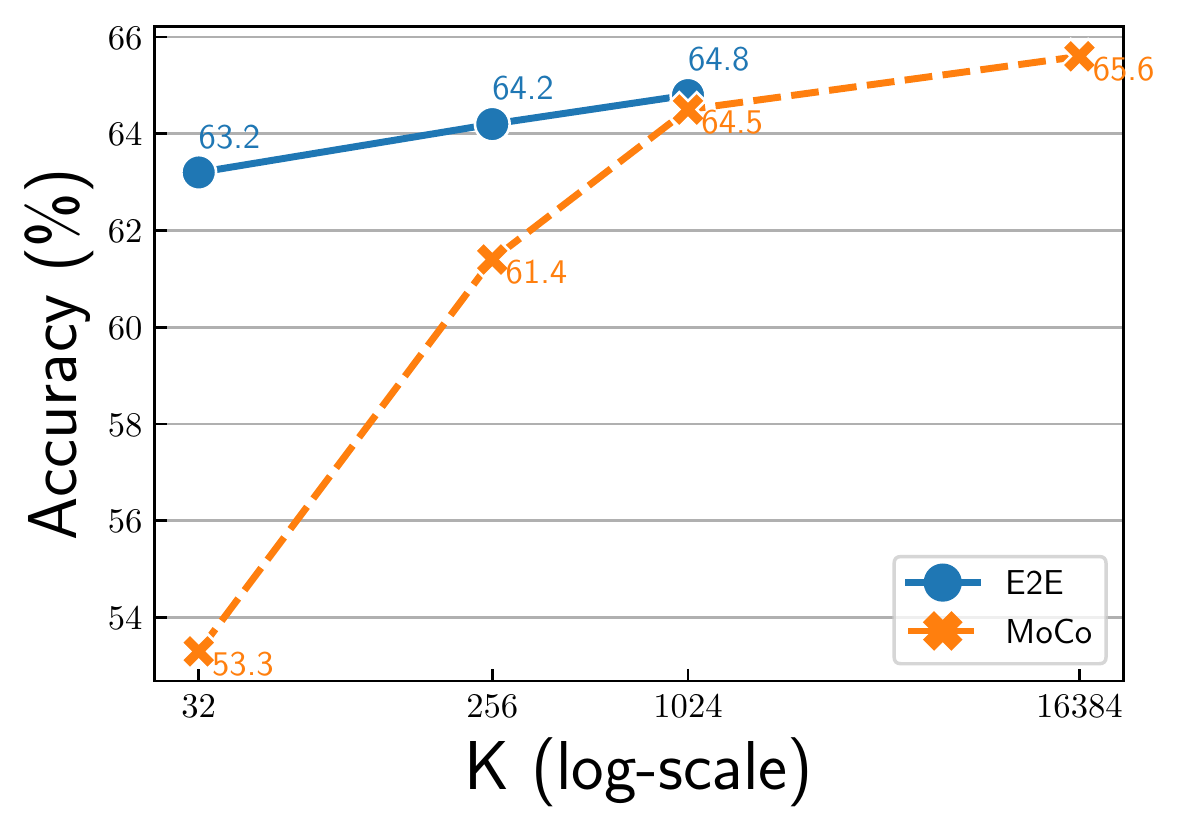}
    }
    \subfloat[COLLAB]{
        \includegraphics[width=0.23\textwidth]{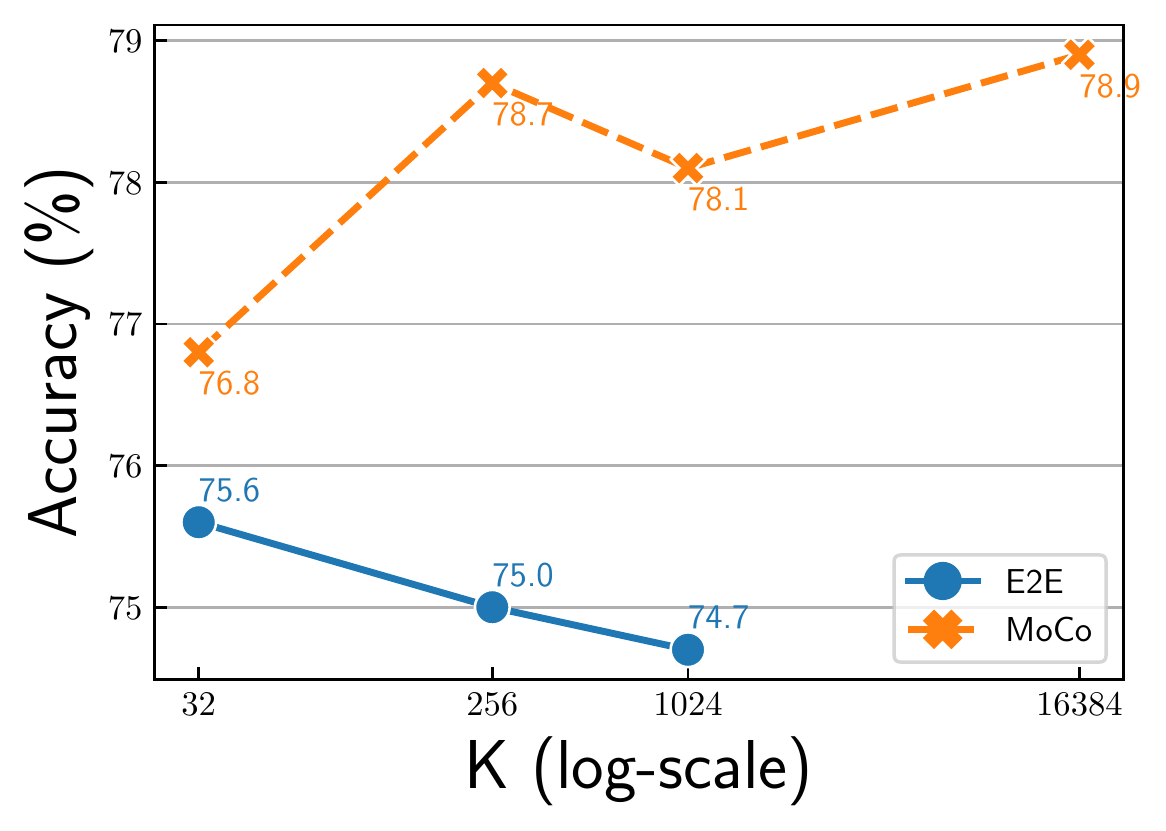}
    }
    \vspace{-0.1in}
    \caption{Comparison of contrastive loss mechanisms.}
    \label{fig:dictionary_size}
    \vspace{-0.1in}
\end{figure}

\vpara{Contrastive loss mechanisms.}
The common belief is that MoCo has stronger expression power than E2E~\cite{he2019momentum},
and a larger dictionary size $K$ always helps.
We also observe such trends, as shown in \Figref{fig:dictionary_size}.
However, the effect of a large dictionary size is
not as significant as reported in computer vision tasks~\cite{he2019momentum}.
For example,
MoCo~($K=16384$) merely outperforms MoCo~($K=1024$) by small margins in terms of accuracy ---
1.0 absolute gain in US-Airport and 0.8 absolute gain in COLLAB.
%We notice that there is a recent theoretical work~\cite{arora2019theoretical} about the role of dictionary size in contrastive learning, which can explain our observation to some extent.
However, training MoCo is much more economical than training E2E.
E2E~($K=1024$) takes 5 days and 16 hours, while MoCo~($K=16384$) only needs 9 hours.
Detailed training time can be found in Table~\ref{fig:time} in the Appendix.

\begin{table}[htbp]
    \caption{Momentum ablation.}
    \vspace{-0.1in}
    \label{tbl:momentum}
    \centering
    \small
    \begin{tabular}{l|rrrrr}
        \toprule
        momentum $m$ & 0    & 0.9  & 0.99 & 0.999         & 0.9999        \\\midrule
        US-Airport   & 62.3 & 63.2 & 63.7 & \textbf{65.6} & 61.5          \\
        COLLAB       & 76.6 & 75.1 & 77.4 & 78.9          & \textbf{79.4} \\
        \bottomrule
    \end{tabular}
    \normalsize
    \vspace{-0.1in}
\end{table}

\vpara{Momentum.} As mentioned in MoCo~\cite{he2019momentum},
momentum $m$ plays a subtle role in learning high-quality representations.
Table~\ref{tbl:momentum} shows  accuracy with different momentum values on US-Airport and COLLAB datasets.
For US-Airport,
the best performance is reached by $m=0.999$,
which is the desired value in \cite{he2019momentum}, showing that building a consistent dictionary is important for MoCo.
However, in COLLAB, it seems that a larger momentum value brings better performance.
Moreover, we do not observe the ``training loss oscillation'' reported in \cite{he2019momentum} when setting  $m=0$.
\shortname{}~(MoCo) converges well, but the accuracy is much worse.

\begin{figure}[htbp]
    \centering
    \subfloat[US-Airport: $y=0.4344x+62.394$]{
        \includegraphics[width=0.23\textwidth]{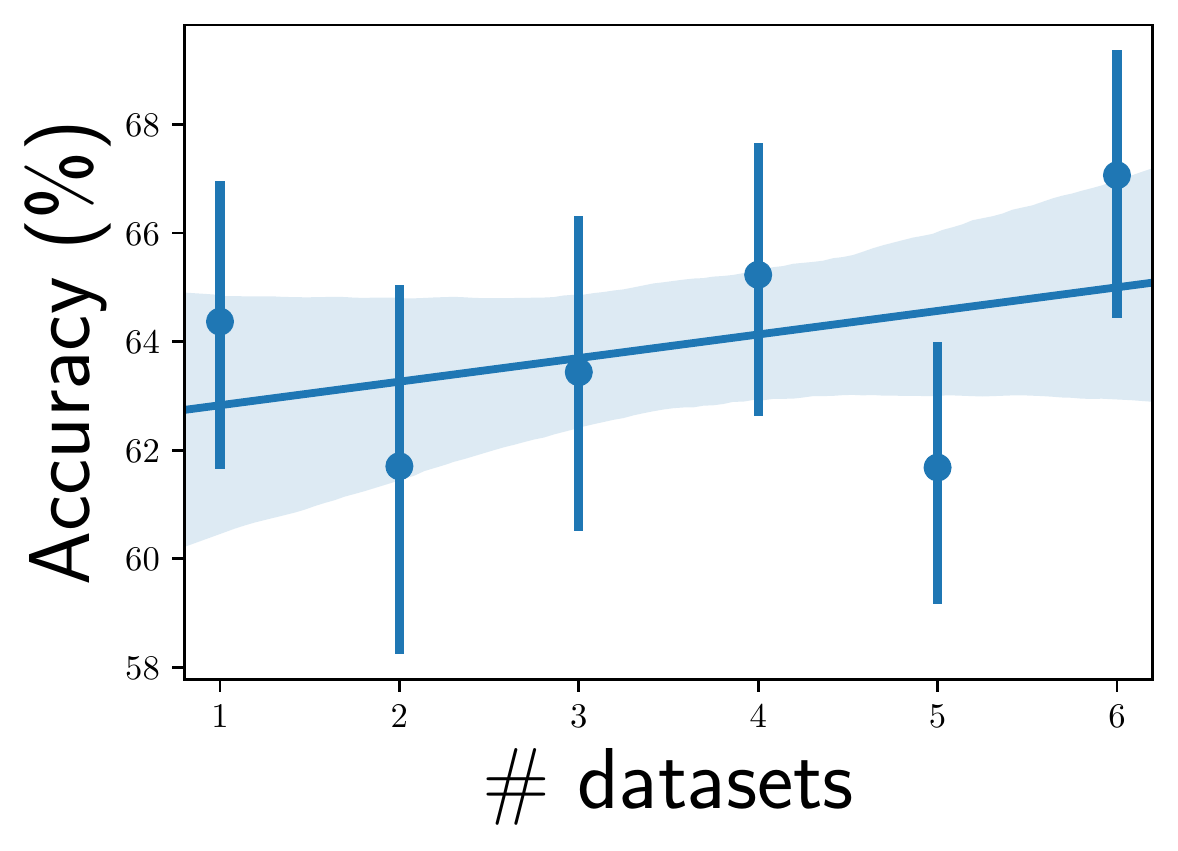}
    }
    \subfloat[COLLAB: $y=0.8065x+74.4737$]{
        \includegraphics[width=0.23\textwidth]{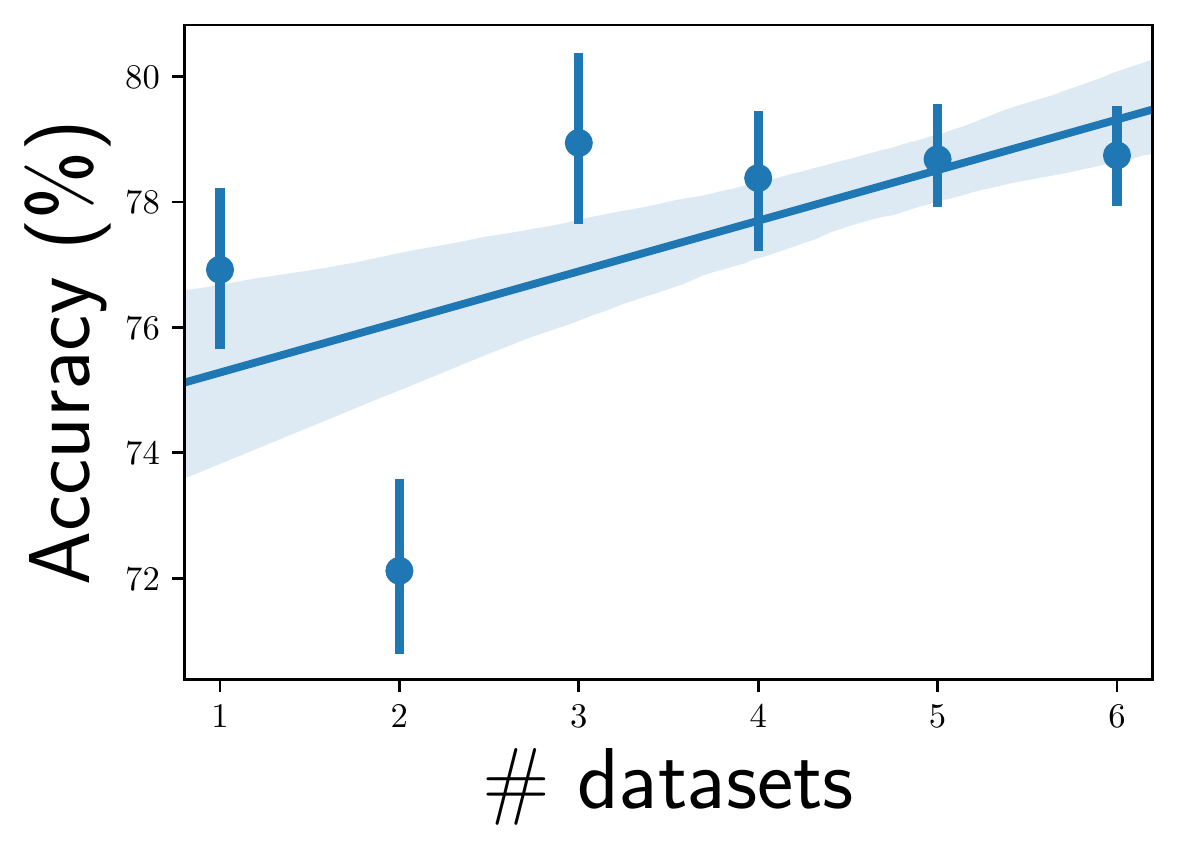}
    }
    \vspace{-0.1in}
    \caption{Ablation study on pre-training datasets.}
    \label{fig:ablation_dataset}
    \vspace{-0.1in}
\end{figure}

\vpara{Pre-training datasets.}
We ablate the number of datasets used for pre-training.
To avoid enumerating a combinatorial space,
we pre-train with first several datasets in Table~\ref{tbl:pretrain_dataset},
and report the 10-fold validation accuracy scores on US-Airport and COLLAB, respectively.
For example, when using one dataset for pre-training, we select Academia; when using two, we choose Academia and DBLP~(SNAP); and so on.
We present ordinary least squares~(OLS) estimates of the relationship between the number of datasets
and the model performance.
As shown in \Figref{fig:ablation_dataset}, we can observe a trend towards higher accuracy when using more datasets for pre-training.
On average, adding one more dataset leads to 0.43 and 0.81 accuracy~(\%) gain on US-Airport
and COLLAB, respectively\footnote{The effect on US-Airport is positive, but statistically insignificant~($p\text{-value}=0.231$),
    while the effect on COLLAB is positive and significant~($p\text{-value}\ll 0.001$).}.

\section{Conclusion}
\label{sec:conclusion}

In this work, we study  
%graph representation learning
the pre-training of 
graph neural networks 
with the goal of characterizing and transferring structural representations in social and information networks.
We present \method{}~(\shortname{}), which 
is a graph-based contrastive learning framework to pre-train graph neural networks from multiple graph datasets. 
The pre-trained graph neural network achieves competitive performance to 
its supervised trained-from-scratch counterparts in three graph learning tasks on ten graph datasets.
In the future, we plan to benchmark more graph learning tasks on more diverse graph datasets, 
%We also would like to explore applications of \shortname{} on graphs in other domains,
such as the protein-protein association networks.%~\cite{szklarczyk2016string}

\vpara{Acknowledgements.}
The work is supported by the
%jie's new 863 on social network
%National High-tech R\&D Program %(2014AA015103,
%sogou 863
%(2015AA124102),
%Guoying Wang
%Development Program of China (2016QY01W0200),
%Maosong's  973
%National Basic Research Program of China (2014CB340506),
%junping du
National Key R\&D Program of China (2018YFB1402600),
%2014CB340500,
%Xiaofeng's 973
%2012CB316006),
%National High Technology Research and Development Program of China (No. 2014AA015103)
%jie's excellent young 
%Natural Science Foundation of China (61222212),
%yuzhou li, wei xu, 
%National Natural Science Foundation of China (61631013, 61532001),
%with wendy
%, 61561130160),
%Maosong Sun's key social science
%National Social Science Foundation of China (13\&ZD190),
NSFC for Distinguished Young Scholar 
%jie tang
(61825602),
%chunyuan zhou
and NSFC (61836013).

%\newpage
\bibliographystyle{ACM-Reference-Format}
%\bibliography{sample-base}
\bibliography{acmart}

\clearpage
\appendix

\section{Appendix}
\begin{table}[htbp]
  \caption{Pre-training hyper-parameters for E2E and MoCo.}
  \label{tbl:hyper}
  \small
  \begin{tabular}{l |r r}
    \toprule
    Hyper-parameter        & E2E    & MoCo   \\\midrule
    Batch size             & 1024   & 32     \\
    Restart probability      & 0.8    & 0.8    \\ 
    Dictionary size $K$    & 1023   & 16384  \\
    Temperature $\tau$     & 0.07   & 0.07   \\
    Momentum $m$           & NA     & 0.999  \\
    Warmup steps           & 7,500  & 7,500  \\
    Weight decay           & 1e-5   & 1e-5   \\
    Training steps         & 75,000 & 75,000 \\
    Initial learning rate  & 0.005  & 0.005  \\
    Learning rate decay    & Linear & Linear \\
    Adam $\epsilon$        & 1e-8   & 1e-8   \\
    Adam $\beta_1$         & 0.9    & 0.9    \\
    Adam $\beta_2$         & 0.999  & 0.999  \\
    Gradient clipping      & 1.0    & 1.0    \\
    Number of layers       & 5      & 5      \\
    Hidden units per layer & 64     & 64     \\
    Dropout rate           & 0.5    & 0.5    \\
    Degree embedding size  & 16     & 16     \\
    \bottomrule
  \end{tabular}
  \normalsize
\end{table}

\begin{figure}[htbp]
  \centering
  \includegraphics[width=0.3\textwidth]{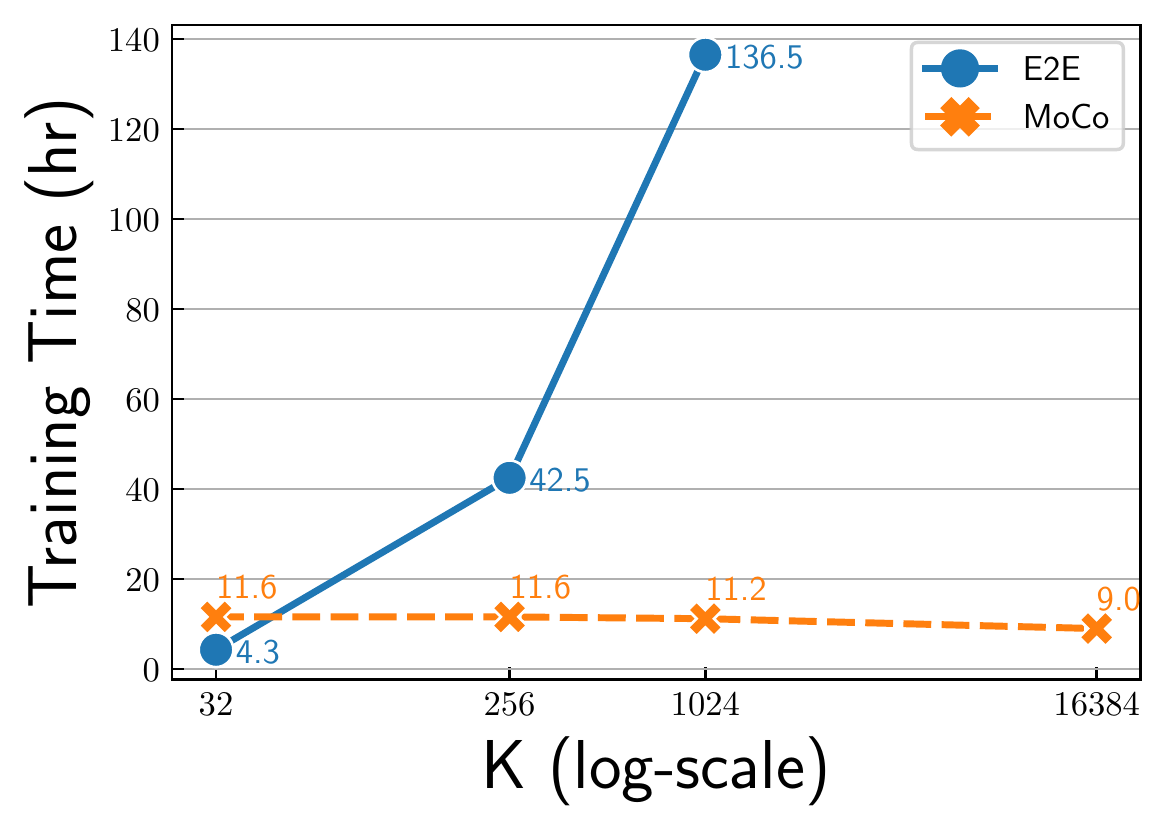}
  \caption{Pre-training time of different contrastive loss mechanisms and dictionary size $K$.}
  \label{fig:time}
\end{figure}

\subsection{Implementation Details}

\subsubsection{Hardware Configuration}\hfill

The experiments are conducted on Linux servers equipped with an Intel(R) Xeon(R) CPU E5-2680 v4 @ 2.40GHz, 256GB RAM and 8 NVIDIA 2080Ti GPUs.

\subsubsection{Software Configuration}\hfill

All models are implemented in PyTorch~\cite{paszke2019pytorch} version 1.3.1, DGL~\cite{wang2019deep} version 0.4.1 with CUDA version 10.1, scikit-learn version 0.20.3 and Python 3.6.
Our code and datasets will be available.

\begin{table}[htbp]
  \caption{Performance of GIN model under various hyper-parameter configurations.}
  \label{tbl:gin}
  \centering
  \small
  \begin{tabular}{@{~}r|@{~}r|@{~}r|@{~}r|@{~}r|@{~}r|@{~}r|@{~}r@{~}}
    \toprule
    batch & dropout & degree & IMDB-B        & IMDB-M        & COLLAB        & RDT-B         & RDT-M         \\\midrule
    32    & 0       & no     & 72.9          & 48.7          & 69.2          & 83.5          & 52.2          \\
    32    & 0.5     & no     & 71.6          & 49.1          & 66.2          & 77.5          & 51.7          \\
    128   & 0       & no     & \textbf{75.6} & 50.9          & 73.0          & \textbf{89.4} & \textbf{54.5} \\
    128   & 0.5     & no     & 74.6          & 51.0          & 72.3          & 81.8          & 53.5          \\
    32    & 0       & yes    & 73.9          & 51.1          & 79.7          & 77.0          & 46.2          \\
    32    & 0.5     & yes    & 74.5          & 50.1          & 79.1          & 77.4          & 45.6          \\
    128   & 0       & yes    & 73.3          & 51.0          & 79.8          & 76.7          & 45.6          \\
    128   & 0.5     & yes    & 73.1          & \textbf{51.5} & \textbf{80.2} & 77.4          & 45.8          \\
    \bottomrule
  \end{tabular}
  \normalsize
\end{table}

\subsubsection{Pre-training}\hfill

The detailed hyper-parameters are listed in Table~\ref{tbl:hyper}.
Training times of \shortname{} variants are listed in \Figref{fig:time}.\footnote{The table shows the elapsed real time for pre-training, which might be affected by other programs running on the server.}
The training time of \shortname{}~(E2E) grows sharply with the dictionary size $K$ while \shortname{}~(MoCo) roughly remains the same, which indicates that MoCo is more economical and easy to scale with larger dictionary size.

\subsection{Baselines}

\subsubsection{Node Classification}\hfill

\vpara{GraphWave~\cite{donnat2018learning}.}
We download the authors' official source code and keep all the training settings as the same.
The implementation requires a networkx graph and time points as input.
We convert our dataset to the networkx format, and use automatic selection of the range of scales provided by the authors.
We set the output embedding dimension to 64.

Code: \url{https://github.com/snap-stanford/graphwave/}.

\vpara{Struc2vec~\cite{ribeiro2017struc2vec}.}
We download the authors' official source code and use default hyper-parameters provided by the authors: (1) walk length = 80; (2) number of walks = 10; (3) window size = 10; (4) number of iterations = 5.

The only modifications we do are: (1) number of dimensions = 64; (2) number of workers = 48 to speed up training.

We find the method hard to scale on the H-index datasets although we set the number of workers to 48, compared to 4 by default.
We keep the code running for 24 hours on the H-index datasets and it failed to finish.
We observed that the sampling strategy in Struc2vec takes up most of the time, as illustrated in the original paper.

Code: \url{https://github.com/leoribeiro/struc2vec}.

\vpara{ProNE~\cite{zhang2019prone}.}
We download the authors' official code and keep hyper-parameters as the same: (1) step = 10; (2) $\theta$ = 0.5; (3) $\mu$ = 0.2.
The dimension size is set to 64.

Code: \url{https://github.com/THUDM/ProNE/}.

\subsubsection{Graph Classification}\hfill

\vpara{DGK~\cite{yanardag2015deep}, graph2vec~\cite{narayanan2017graph2vec}, InfoGraph~\cite{sun2019infograph}, DGCNN~\cite{zhang2018end}.}
We adopt the reported results in these papers.
Our experimental setting is exactly the same except for the dimension size.
Note that graph2vec uses 1024 and InfoGraph uses 512 as the dimension size.
Following GIN, we use 64.

\vpara{GIN~\cite{xu2018how}.}
We use the official code released by \cite{xu2018how} and follow exactly the procedure described in their paper:
the hyper-parameters tuned for each dataset are:
(1) the number of hidden units $\in \{16, 32\}$ for bioinformatics graphs and $64$ for social graphs;
(2) the batch size $\in \{32, 128\}$;
(3) the dropout ratio $\in \{0, 0.5\}$ after the dense layer;
(4) the number of epochs, i.e., a single epoch with the best cross-validation accuracy averaged over the 10 folds was selected.
We report the obtained results in Table~\ref{tbl:gin}.

Code: \url{https://github.com/weihua916/powerful-gnns}.

\subsubsection{Top-$k$ Similarity Search}\hfill

\vpara{Random, RolX~\cite{henderson2012rolx}, Panther++~\cite{zhang2015panther}.}
We obtain the experimental results for these baselines from \citet{zhang2015panther}.

Code: \url{https://github.com/yuikns/panther/}.

\vpara{GraphWave~\cite{donnat2018learning}.}
Embeddings computed by the GraphWave method also have the ability to generalize across graphs.
The authors evaluated on synthetic graphs in their paper which are not publicly available.
To compare with GraphWave on the co-author datasets, we compute GraphWave embeddings given two graphs $G_{1} \text { and } G_{2}$ and follow the same procedure mentioned in \secref{subsubsec:similarity} to compute the HITS@10~(top-10 accuracy) score.

Code: \url{https://github.com/snap-stanford/graphwave/}.

\subsection{Datasets}

\subsubsection{Node Classification Datasets}\hfill

\vpara{US-Airport}\footnote{\url{https://github.com/leoribeiro/struc2vec/tree/master/graph}.}
We obtain the US-Airport dataset directly from \citet{ribeiro2017struc2vec}.

\vpara{H-index}\footnote{\url{https://www.openacademic.ai/oag/}.}
We create the H-index dataset, a co-authorship graph extracted from OAG~\cite{zhang2019oag}.
Since the original OAG co-authorship graph has millions of nodes, it is too large as a node classification benchmark.
Therefore, we implemented the following procedure to extract smaller subgraphs from OAG:

% \jz{I decide to keep H-index3 dataset only. BTW, can we rename this dataset to OAG?}

\begin{enumerate}[leftmargin=*,itemsep=0pt,parsep=0.5em,topsep=0.3em,partopsep=0.3em]
  \item Select an initial vertex set $V_s$ in OAG;
  \item Run breadth first search (BFS) from $V_s$ until $N$ nodes are visited;
  \item Return the sub-graph induced by the visited $N$ nodes.
\end{enumerate}

We set $N = 5,000$, and randomly select 20 nodes from top 200 nodes with largest degree as he initial vertex set in step (1).
% and use different strategies to select 
% For H-index1, we select the node with the largest degree;
% For H-index2, we randomly select a node in the graph;

\subsubsection{Graph Classification Datasets}
\footnote{\url{https://ls11-www.cs.tu-dortmund.de/staff/morris/graphkerneldatasets}}
\hfill

We download COLLAB, IMDB-BINARY, IMDB-MULTI, REDDIT-BINARY and REDDIT-MULTI5K from Benchmark Data Sets for Graph Kernels~\cite{kersting2016benchmark}.

\subsubsection{Top-$k$ Similarity Search Datasets}
\footnote{\url{https://github.com/yuikns/panther}}
\hfill

We obtain the paired conference co-author datasets, including KDD-ICDM, SIGIR-CIKM, SIGMOD-ICDE, from the \citet{zhang2015panther} and make them publicly available with the permission of the original authors.

\end{document}